\def\year{2020}\relax
\documentclass[letterpaper]{article} 
\usepackage{aaai20}  
\usepackage{times}  
\usepackage{helvet} 
\usepackage{courier}  
\usepackage[hyphens]{url}  
\usepackage{graphicx} 
\usepackage{amsmath}
\usepackage{amssymb}
\usepackage{subfigure}
\usepackage{graphicx}  
\title{KPNet: Towards Minimal Face Detector}
\author{Guanglu Song\textsuperscript{\rm 1,3}\ \ \  Yu Liu\textsuperscript{\rm 2}\thanks{Corresponding author}\ \ \  Yuhang Zang\textsuperscript{\rm 1}\ \ \ Xiaogang Wang\textsuperscript{\rm 2}\ \ \ Biao Leng\textsuperscript{\rm 3,4}\ \ \  Qingsheng Yuan\textsuperscript{\rm 5} \\
\textsuperscript{\rm 1}SenseTime X-Lab\\
\textsuperscript{\rm 2}The Chinese University of Hong Kong, Hong Kong\\
\textsuperscript{\rm 3}School of Computer Science and Engineering, Beihang University, Beijing 100191, China\\
\textsuperscript{\rm 4}Beijing Advanced Innovation Center for Big Data and Brain Computing, Beihang University, Beijing, 100191\\
\textsuperscript{\rm 5}National Computer network Emergency Response technical Team/Coordination Center of China \\
\textsuperscript{\rm 1}\{songguanglu, zangyuhang\}@sensetime.com, \textsuperscript{\rm 2}\{yuliu, xgwang\}@ee.cuhk.edu.hk, \textsuperscript{\rm 3}lengbiao@buaa.edu.cn,
\textsuperscript{\rm 5}yqs@cert.org.cn
}
 
\begin{document}

\maketitle

\begin{abstract}
The small receptive field and capacity of minimal neural networks limit their performance when using them to be the backbone of detectors. In this work, we find that the appearance feature of a generic face is discriminative enough for a tiny and shallow neural network to verify from the background. And the essential barriers behind us are 1) the vague definition of the face bounding box and 2) tricky design of anchor-boxes or receptive field. Unlike most top-down methods for joint face detection and alignment,
the proposed KPNet detects small facial keypoints instead of the whole face by in the bottom-up manner.
It first predicts the facial landmarks from a low-resolution image via the well-designed fine-grained scale approximation and scale adaptive soft-argmax operator.
Finally, the precise face bounding boxes, no matter how we define it, can be inferred from the keypoints. Without any complex head architecture or meticulous network designing, the KPNet achieves state-of-the-art accuracy on generic face detection and alignment benchmarks with only $\sim1M$ parameters, which runs at 1000fps on GPU and is easy to perform real-time on most modern front-end chips.
\end{abstract}

\section{Introduction}\label{sec:introduction}
\begin{figure}[t]
\centering
\includegraphics[width=0.8\linewidth]{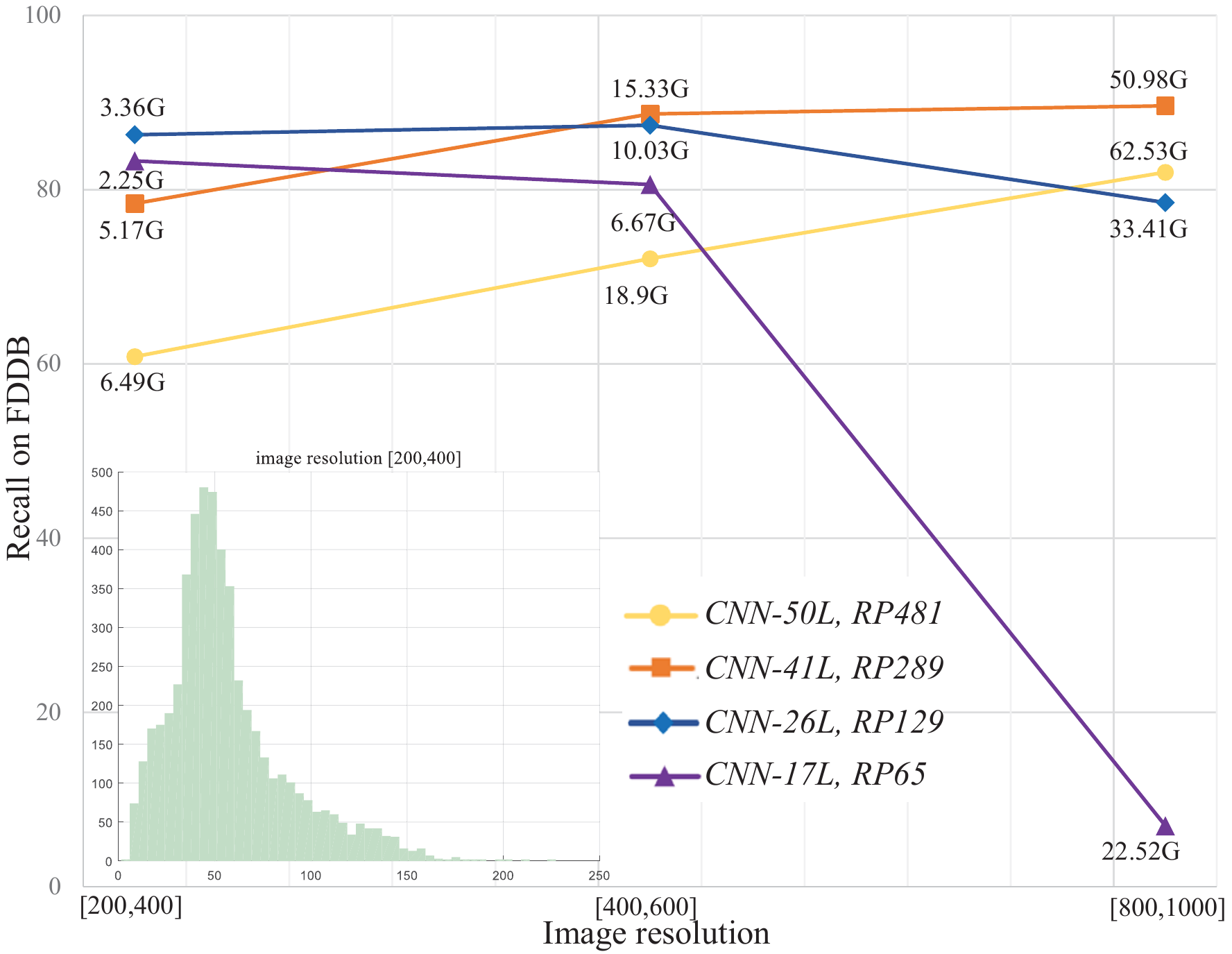}
   \caption{Performance of different networks with variant input resolutions on FDDB. $RP \#$ means the receptive field and the numbers along with the points represent the multiply-add operations. The distribution diagram in the lower-left corner represents the scale distribution of ground truth boxes when the input image resolution is $[200,400]$.}
\label{fig:intro}
\end{figure}

The performance of face detection has been constantly improved thanks to the anchor-based mechanism~\cite{girshick2015fast,ren2015faster} with the top-down strategy.
By simply assigning dense anchor templates in complex models, we can obtain a face detector with excellent performance.
In the current state-of-the-art research, the essence of face detection is how to design the receptive field adaptive to large scale-variance.
With the emergence of some seminal works~\cite{hu2017finding,zhang2017s3fd,najibi2017ssh,zeng2017crafting,li2019gradient} to explore the relationship between the receptive field and the large scale-variance, the performance of face detection is further improved.
Inspired by this, the FPN-style framework~\cite{lin2017feature,li2019zoom} has become a priority choice for researchers and it can effectively enhance the performance of face detectors to handle faces with different scales.
Encouraged by these insights, most of the state-of-the-art algorithms~\cite{tang2018pyramidbox,chi2018selective,li2018dsfd} construct adaptive receptive fields to detect faces. They either detect targets with different scales at different levels of the network or detect targets by fusing enhanced features generated by different levels.
With the assistance of complex deep backbone~\cite{he2016deep}, improving face detection performance with refined receptive field~\cite{zhang2017s3fd,najibi2017ssh} or embedding new enhanced modules~\cite{tang2018pyramidbox,chi2018selective,li2018dsfd} has become the guidance in the field of face detection.
However, these top-down approaches with complex backbone networks lead to a heavy computational burden,
even though some novel works~\cite{song2018beyond,liu2017recurrent} are proposed to accelerate them.
Under the constraints of these current mechanisms, we naturally raise a question: \textbf{can large scale-variance be solved only through a deeper and more complex backbone with well-designed strategies?}

Keep the bottleneck of current research in mind, this paper tries to seek the answer to this question.
We re-explore face detection from the two aforementioned essential factors: the receptive field and the large scale-variance. 
The above question is decomposed into the following more detailed sub-problems and different controlled experiments are performed on FDDB to seek answers.
\begin{itemize}
\item It's a common practice to up-sample the images to $480\times640$ or even $800\times1000$ which facilitates better performance based on a complex backbone network with a large receptive field. Will the lightweight network fail in this configuration?
\item How do complex and lightweight networks with different receptive fields perform on low-resolution input images?
\end{itemize}

We design detection backbones of various depths with different receptive fields by modifying ResNet50~\cite{he2016deep}.
Four detectors CNN-50L, CNN-41L, CNN-26L and CNN-17L with depth 50, 41, 26 and 17 are performed on FDDB. 
The training set is same with ~\cite{liu2017recurrent} and the recall of top 100 proposals for each image is used for evaluation. Results are shown in Fig.~\ref{fig:intro} and the anchor setting is that $A=\{[16\sqrt{2}, 16\sqrt{2}], [32\sqrt{2}, 32\sqrt{2}], [64\sqrt{2}, 64\sqrt{2}]\}$ for image resolution [200, 400] to detect face scale [16, 128],  $2A$ for image resolution [400, 600] to detect face scale [32, 256] and $4A$ for image resolution [800, 1000] to detect face scale [64, 512].
According to the results, the former questions can be explained. 
When the lightweight network adopts the high-resolution image, even if the appropriate anchor templates are assigned, it still fails in this configuration due to the limitation of the receptive field. 
It's worth noting that \textbf{shrinking the image to low resolution with lightweight backbone can still lead to the comparable performance to the deeper and more complex backbone.}

However, potential barriers still exist behind this discovery.
The accurate face boxes heavily rely on the tricky design of anchor boxes or receptive field and also, the vague definition of face boxes (e.g. face boxes in FDDB are defined by the ellipse) can easily degrade the performance.
Fortunately, bottom-up methods can effectively get rid of the bottleneck in top-down mechanism via converting boxes to keypoints~\cite{law2018cornernet,zhou2019bottom}.
So these ideas naturally lead to a simple, lightweight but accurate framework KPNet
where two essential factors are embedded into it, one is to shrink the input image to low resolution with lightweight backbone and the other is to shrink the face concept from box to keypoints to skip the deficiency of general pipeline.

Beyond the general face detection pipeline, the precise facial keypoints can be located first via the carefully designed algorithm and then the accurate face boxes can be inferred by it.
So that it can perform as a bottom-up approach to joint face detection and alignment.
Different from other top-down algorithms~\cite{zhang2016joint,king2009dlib} for joint face detection and alignment,
KPNet bypasses the vague definition of bounding boxes and takes advantage of the less uncertainty definition of keypoints.
Moreover, it's different from the bottom-up approaches in pose estimation where each landmark is independently predicted and associative embedding~\cite{newell2017associative,law2018cornernet} is used to group them into an instance.
The well-designed fine-grained scale approximation in KPNet can potentially imply the group of landmarks and with the scale adaptive soft-argmax, it can straightforwardly predict the landmarks belonging to the same face.

To summarize, the contributions of this paper are as follows:

1) According to the re-exploration on face detection and the advantages of shrinking target concept from box to keypoints, we propose a novel KPNet with the simple, lightweight but accurate mechanism for the generic face ($>$20 px) detection and alignment.

2) We propose the fine-grained scale approximation and scale-based customization soft-argmax operator to improve the performance by a large margin.

3) Different from all of the joint face detection and alignment methods that adopt the top-down pipeline, KPNet
follows the bottom-up mechanism and the more precise definition of landmarks than boxes enables the better performance.

4) Without bells and whistles, KPNet can achieve the SOTA performance on generic face detection benchmarks FDDB, AFW, MALF, and face alignment benchmark AFLW. And also the model inference speed with the offline application can achieve $\sim1000$fps on GTX 1080Ti.

\section{Related Works}

\begin{figure*}
\centering
\includegraphics[width=0.85\linewidth]{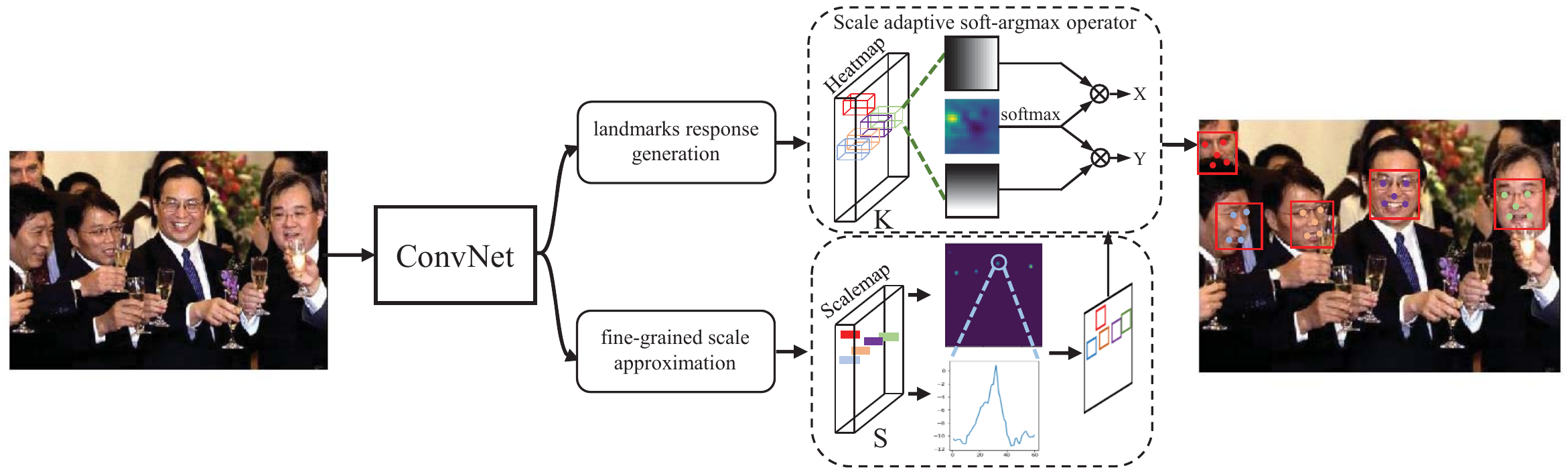}
   \caption{Overview of KPNet. The backbone is followed by two specific modules, one for generating the landmarks response map and the other for approximating the fine-grained face scale. Utilizing the predictions from both modules,  the landmarks extractor locates the facial keypoints and infer the face boxes. $K$ and $S$ represent the channel numbers of the feature maps.}
\label{fig:pipe}
\end{figure*}

{\bf{Face detection.}}
Since the emergence of the powerful CNN~\cite{he2016deep}, the performance of face detection has been improved by a large margin.  
With the success of anchor-based methods such as Fast RCNN~\cite{girshick2015fast} and Faster RCNN~\cite{ren2015faster} on object detection,
several different approaches~\cite{wang2017face,wang2017detecting} are inspired by them and achieve satisfactory performance on face detection.
More recently, the FPN-style framework~\cite{lin2017feature} encourages the researchers to explore the relationship between the receptive field of the face detector and the anchor design skills~\cite{zhang2017s3fd}.
Benefiting from these explorations, detecting faces with different scales from different layers in a single network~\cite{najibi2017ssh,tang2018pyramidbox,yang2017face} has determined its position in the field of face detection.
Several works~\cite{li2018dsfd,chi2018selective} detect a face from the feature fusion of different layers and the enhanced feature make it robust for scale-variance.
The deeper and complex backbone with the FPN-style framework achieves the new state-of-the-art and this idea of detector design has dominated face detection for many years. 
Unfortunately, the accurate face detection heavily relies on the tricky design of anchor boxes or receptive field. 
Moreover, the vague definition of face boxes makes it hard to generalize to generic face detection.

{\bf{Face alignment.}}
Face alignment refers to facial landmark detection and it mainly focuses on identifying the geometry structure of the human face. The CNN-based face alignment methods can be divided into two categories, i.e., coordinate regression model and heatmap regression model.

The coordinate regression model directly regresses the facial landmarks from the input image. Many works~\cite{dong2018supervision,feng2018wing} have the advantage of explicit inference of landmarks without any post-processing.  However, these regression-based methods are not performing as well as heatmap regression models.
The heatmap regression models generate likelihood heatmaps for each keypoint, respectively. 
In these heatmap-based methods, hourglass networks~\cite{newell2016stacked} becomes the backbone of many works~\cite{newell2016stacked,deng2017joint} due to its capabilities of obtaining multi-scale information. Some works~\cite{lv2017deep,sun2013deep} have adopted facial parts to aid face alignments tasks and LAB~\cite{wu2018look} uses more precise facial boundary to assist in detecting facial landmarks.

{\bf{Joint face detection and alignment.}}
Face alignment and face detection are closely related, yet few works~\cite{zhang2016joint,king2009dlib} jointly perform them.
The popular algorithm MTCNN~\cite{zhang2016joint} follows the general top-down mechanism which first regresses the face boxes and then generates the corresponding landmarks. 
However, the accurate face boxes generation relies on the tricky design of anchor boxes and is weak against the vague definition of face boxes.
Comparing with it, KPNet first predicts the facial landmarks with unambiguous definition and then inference the face boxes by it.
The more precise definition of landmarks than face boxes enables it to achieve better performance than top-down methods.

\section{KPNet}
\subsection{Overview}
In KPNet, we adopt the bottom-up mechanism to perform face detection and alignment simultaneously.
Fig.~\ref{fig:pipe} provides an overview of KPNet. 
Firstly, the potential face scale proposals can be predicted by the fine-grained scale approximation. Secondly, the keypoints can be computed by the scale adaptive soft-argmax with the scalemap and the output heatmap of landmarks response generation. 
Finally, we apply a simple transformation algorithm~\cite{song2018beyond} to infer the final bounding boxes from the landmarks.

\subsection{Fine-grained scale approximation}
To better detect the facial keypoints, we need to locate the focus regions where existing faces are.
The anchor-based mechanism is undoubtedly an appropriate solution,
but its' sophisticated design techniques make it a departure from the simplicity and flexibility of our framework.
Inspired by~\cite{hao2017scale,liu2017recurrent} where CNN is capable of approximating the scale information in the low-resolution image,
we convert the boxes regression to fine-grained face scale classification for each pixel in a feature map.
Different from ~\cite{liu2017recurrent} where only the existing scales are predicted to select valid layers from the image pyramid,
in KPNet, we add additional spatial information to scale approximation.

The scalemap is generated by fine-grained scale approximation that consists of only one convolutional layer with kernel size 3 is used.
It is a probability map $M$ with dimension $H'\times W'\times S$, where $S$ is the predefined 
number of scales.
Given an image $I$ with face boxes $[x, y,h,w]$ where $(x,y)$ means the face center and $h,w$ represent the height and width of the face,
the $M$ is firstly initialized to 0 and then we calculate the active channel index $b$ as:
\begin{eqnarray}
\begin{split}
b = 10\times(log_2{\frac{max(h,w)\times2048}{I_{max}}}-5) \label{eq:scale},
\end{split}
\end{eqnarray}
where b is the index from 1 to $S$ and $I_{max}$ represents the max edge of $I$.
The minimum detected face size for $I_{max}$=1280 in 720P image is $\bf{20px}$.
In Eq.~\ref{eq:scale}, the face size from $2^5$ to $2^{11}$ can be mapped into the different channel indexes in $M$. For simplicity, we divide the $[2^t,2^{t+1}]$, $t\in[5,10]$ into 10 scale bins and thus the total scale number $S=60$.
With the computed $b$, the value at coordinate $(\lfloor \frac{x}{N_s} \rfloor, \lfloor \frac{y}{N_s} \rfloor,b)$ can be defined as:
\begin{eqnarray}
\begin{split}
M(\lfloor \frac{x}{N_s} \rfloor, \lfloor \frac{y}{N_s} \rfloor,b) = 1 ,
\end{split}
\end{eqnarray}
where $N_s$ means the stride of the network.
Encouraged by~\cite{wu2018look,law2018cornernet}, to alleviate the difficulty of feature learning in the discrete distribution,
we introduce the 2D Gaussian function to refine it.
Given the radius $r=\lfloor \frac{b}{10} \rfloor$ and the host point $(x_h,y_h)=(\lfloor \frac{x}{N_s} \rfloor, \lfloor \frac{y}{N_s} \rfloor)$, the values of its neighbouring points can be formulated as:
\begin{eqnarray}
\begin{split}
M(x_i,y_i,b) = e^{\frac{(\frac{x_i-x_h}{r})^2+ (\frac{y_i-y_h}{r})^2  }{2\sigma^2}} ,
\end{split}
\end{eqnarray}
 where $(x_i,y_i)$ belongs to the neighbour set $\mathcal{N}(x_h,y_h,b)$ and $\sigma$ is set to 0.1 in our experiments.
 During training, the input image is resized with the higher dimension equal to 256 and the loss of the scalemap training is a binary multi-class cross entropy loss:
 \begin{eqnarray}
\begin{split}
L_{scale} = -\frac{1}{|M|}\sum^{|M|}_{i=0} (p_i log{\hat{p}_i} +(1-p_i)log(1-\hat{p_i}) )\label{bceloss} ,
\end{split}
\end{eqnarray}
where $p_i,\hat{p_i}$ are the ground truth label and prediction of the i-th pixel in $M$. $|M|$ indicates the pixel number in feature map $M$.

For inference, given a threshold, we select all of the valid coordinates $(x_v, y_v, b_v)$ from $M$ and compute the scale $s_v=max(h, w)$ by Eq.~\ref{eq:scale} based on its channel index $b_v$.
Finally, the scale proposal $[x_v, y_v,s_v, s_v]$ can be obtained.

\subsection{Scale adaptive soft-argmax operator}
As shown in~\cite{wu2018look}, heatmap regression models can achieve better performance than coordinate regression models do.
With this conclusion in mind, instead of regressing the keypoint coordinates, we detect them from the landmark response map.
At the end of the backbone, the landmark response generation is used to generate a response map with dimension $H\times W\times K$ where K is the number of facial keypoints.
It only consists of one convolutional layer with kernel size 3.
Instead of the argmax function, which is not differentiable, breaking the learning chain on neural networks~\cite{luvizon2017human}, 
we propose the $scale$ $adaptive$ $soft$-$argmax$ operator which keeps the properties of specialized part detectors while being fully differentiable.
Different from the usage in~\cite{luvizon2017human} where it applies to the global response map cooperated with top-down methods,
the proposed $scale$ $adaptive$ $soft$-$argmax$ is performed on the scale aware locations cooperated with the bottom-up pipeline.

Given a scale proposal $\mathcal{S}$$=$$[x_1,y_1,x_2,y_2]$ where $x_1,y_1,x_2,y_2$ mean the top left and bottom right corner, we define the $\emph{Softmax}$ operation on it $h\in \mathbb{R}^{H\times W\times K}$
as:

\begin{equation}
\label{5}
\Phi(h_{i,j,c})=\left\{
\begin{aligned}
 \frac{e^{h_{i,j,c}}}{\sum_{m=x_1}^{x_2}\sum_{n=y_1}^{y_2}e^{h_{m,n,c}}} & , & (i,j)\in\mathcal{S}, \\
0 & , & others.
\end{aligned}
\right.
\end{equation}

where $h_{i,j,c}$ is the value of heat map $h$ at location $(i,j)$ of channel $c$.
The coordinates of the landmarks $P_c= (\Psi_{c,x},\Psi_{c,y})$ corresponding to the $\mathcal{S}$ 
are given by:
 \begin{eqnarray}
\begin{split}
\Psi_{c,x} = \sum_{i=1}^{W}\sum_{j=1}^{H}\frac{\mathbb{P}(i-x_1,w)}{w}\Phi(h_{i,j,c}) \\
\Psi_{c,y} = \sum_{i=1}^{W}\sum_{j=1}^{H}\frac{\mathbb{P}(j-y_1,h)}{h}\Phi(h_{i,j,c}), \label{6}
\end{split}
\end{eqnarray}
where $w,h$ indicate the width and height of $\mathcal{S}$. $\mathbb{P}(x,y)$ returns $x$ when $0\leq x \leq y$ and 0 otherwise.

For keypoints regression based on each $h$, we adopt the $L_{keypoint}$ loss function as follows:
 \begin{eqnarray}
\begin{split}
L_{keypoint} = \frac{1}{2K}\sum_{c=1}^{K}\left \| \Psi_{c,*} - \mathcal{G}_{c,*} \right \|^2_2 ,
\end{split}
\end{eqnarray}
where $ \mathcal{G}_{c,*}$ means the ground truth keypoints.
After obtaining the facial keypoints, the face boxes can be inferred by it conforming to the definition of us.
Define the probability of a scale proposal as $P_s$, the score of the corresponding face box is formulated as:
 \begin{eqnarray}
\begin{split}
P = P_s + \sum_{c=1}^{3}max(\Phi(h_{*,*,c})) ,
\end{split}
\end{eqnarray}
where $max(\cdot)$ means the maximum operation on the spatial resolution and c from 1 to 3 represents the channels corresponding to
left eye, right eye, and nose. We empirically utilize the keypoints information to weaken some false positive scale proposals.
\begin{figure}
\centering
\includegraphics[width=0.35\linewidth]{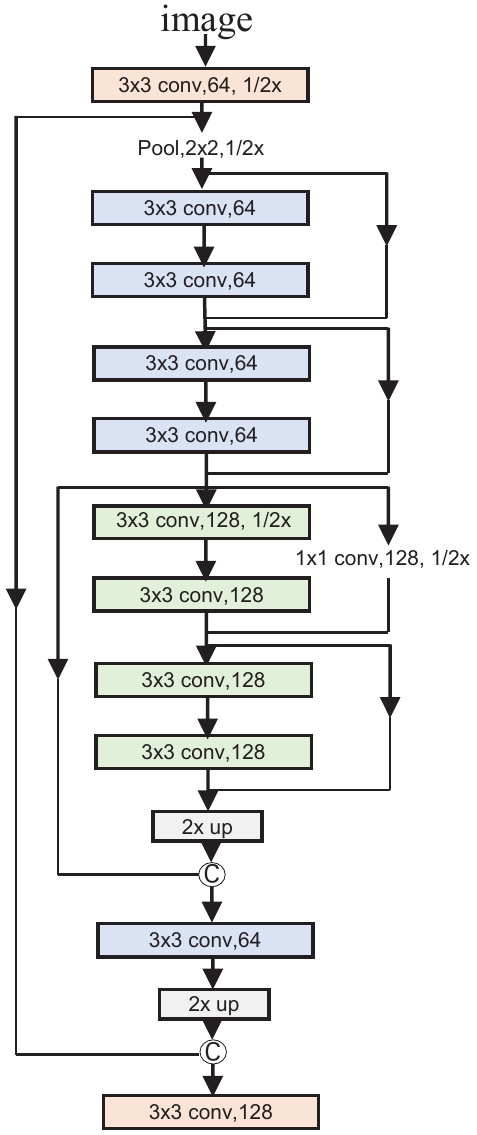}
   \caption{Details of DRNet. `C' means the concatenate operation and other skip operations represent element-wise sum.}
\label{fig:arch}
\end{figure}

\subsection{Backbone architecture}\label{arch}
We use two different CNN architectures as the backbones of KPNet, respectively. One is the robust stacked hourglass network and the other is 
DRNet with faster inference speed.
The hourglass network only consists of one stacked hourglass and the channel number is reduced to 64. The first convolution layer with kernel size $7\times 7$, stride 2 is replaced by two small convolution layers without BN and ReLU between them. Kernel size $3\times 3$ with stride 2 and kernel size $3\times3$ with stride 1 are assigned to them, respectively. Furthermore, we upsample the spatial resolution by a factor of 2 (using nearest neighbor upsampling for simplicity) at the end of the hourglass to reduce the bias caused by heavy down-sample operations. 
Finally, the total stride of the hourglass is 2 and the parameter is $\sim1.04M$.
Even though the network has become very lightweight after these specific modifications, 
abundant operations on large resolution feature maps still limit the inference speed of the network.

Following the principle of reducing the redundant operation on high-resolution feature maps, we design a simple and fast De-redundancy Net (DRNet).
It can achieve $\sim5\times$ faster inference speed than the former hourglass with the same number of parameters.
The details of DRNet are shown in Fig~\ref{fig:arch}.
DRNet is a simple and lightweight network with only 11 layers.
Given an input image, we first reduce it $4\times$ via a $3\times 3$ convolution layer with stride 2 and a $2\times 2$ max pooling layer with stride 2. Then some residual blocks are followed and two nearest neighbor upsampling layers with factor 2 are used to upsample the spatial resolution. The total stride of DRNet is 2 and the large-span skip layer enables the network to retain as much input information as possible which makes it comfortable for low-resolution input. The lightweight structure without redundant operations on high-resolution feature maps ensures that it can achieve the faster inference speed.

\subsection{Advantage insights of KPNet}
All of the joint face detection and alignment algorithms~\cite{zhang2016joint,king2009dlib} first generate the
face boxes and then predict the corresponding landmarks.
Limited by face detection accuracy which heavily relies on the tricky design of anchor boxes and is easily influenced by the ambiguous definition of bounding boxes, it's hard to achieve better performance with faster speed.
Compared with these top-down pipelines, KPNet has the following advantages.
First of all, KPNet adopts the bottom-up mechanism that first locates the landmarks with the unambiguous definition, and then the face boxes can be inferred from keypoints. The precise definition of landmarks compared with face boxes allows it to easily achieve higher performance.
Secondly, KPNet skips anchor designing and thusly is flexible to deploy this network without complicated design skills.
Finally, different from the most face detection methods~\cite{li2018dsfd,liu2017recurrent}, KPNet does not depend on high-resolution input. Through the combination of low-resolution input and lightweight network, it can achieve SOTA performance on both of the generic face detection ($>$20px) and alignment with fast inference speed (offline application with $\sim1000$fps).

\section{Experiments}
\subsection{Implement details}
We implement KPNet in PyTorch. Both of the hourglass and DRNet are randomly initialized under the default setting of PyTorch without pretraining on any external dataset.
During training, we set the input resolution of the network to $256\times 256$, which leads to an output resolution of $128\times 128$.
For training on generic face detection, we adopt the training set the same as~\cite{liu2017recurrent} and none of the data augmentations is performed. For joint face detection and alignment, $K$ is set to 5 representing the left eye, right eye, nose, left corner of the mouth and right corner of the mouth.
We joint optimize the loss function $L_{scale}$ and the $L_{keypoint}$ with lossweight 1:1 via SGD.
Due to the huge pixels in scalemap $M\in \mathbb{R}^{128\times 128\times 60}$, the weight of $L_{scale}$ is set to 10000 for faster convergence.
Benefiting from the low-resolution input and lightweight backbone, we use a batch size of 128 and train the network on 4 GTX 1080Ti GPUs.
For training on AFLW, because of the missing annotation of some face boxes in the training set, we only train the $L_{keypoint}$ for the annotated facial landmarks. $K$ is set to 19 to correspond to the annotated facial landmarks in AFLW.
We adopt the data augmentation strategy the same as~\cite{feng2018wing} for preventing over-fitting.
We train the network for 150k iterations with a learning rate warmup strategy.
The learning rate is linearly increased to 0.01 from 0.00001 in the first 50k iterations and we reduce it to 0.001 for the last 50k iterations.

At the inference stage, we first generate the scale proposals through the predefined threshold from scalemap, and then compute the corresponding keypoints via then scale adaptive soft-argmax according to Eq.~\ref{5} and Eq.~\ref{6}.
Finally, NMS with IOU 0.6 is adopted on the face boxes inferred from these keypoints.

\subsection{Test benchmarks}
We evaluate KPNet on the generic face detection benchmarks FDDB~\cite{jain2010fddb}, AFW, MALF, and face alignment benchmark AFLW~\cite{koestinger2011annotated}.
We adopt the relabeled version provided by~\cite{liu2017recurrent} where some missing faces are re-annotated.
We follow~\cite{feng2018wing} to adopt the AFLW-Full in our experiments where 20,000 and 4,386 images are used for training and testing, respectively.
For face detection, we follow the protocol of~\cite{liu2017recurrent}.

\begin{table}[h]
\centering
\begin{center}
\begin{tabular}{c|c|c|c}
\hline  
Network & FDDB (\%)& AFW (\%) & MALF (\%)\\
\hline
RSA$_{base}$ & 96.0 & 100.0 & 96.49\\
DRNet & 96.6 & 99.6 & 97.1\\
Hourglass$_{light}$ & 96.7& 99.8 & 97.59\\
\hline 
\end{tabular} 
\end{center}
\caption{ Recall on FDDB, AFW, and MALF. All of the results are evaluated on the top 100 proposals.}
\label{tab:scale_recall}
\end{table}
\subsection{Ablation study}
{\bf{Fine-grained scale approximation.}} Fine-grained scale approximation is a key component of KPNet. To understand its performance, 
we directly evaluate its recall on FDDB, AFW, and MALF. 
Similar to the evaluation metric in former section, we compute the recall of the top 100 scale proposals. Furthermore, we implement the anchor-based detector RSA$_{base}~\cite{liu2017recurrent}$, which is the SOTA algorithm on generic face detection. According to their claimed configuration, we evaluate the recall for comparison with the same protocol.  
The result is shown in Tab.~\ref{tab:scale_recall}. Hourglass$_{light}$ is the modified hourglass network in Sec.~\ref{arch}
The fine-grained scale approximation can achieve a comparable recall to the SOTA anchor-based algorithm without relying on the experience design.

{\bf{Advantage of scale adaptive soft-argmax operator.}}
We introduce the scale adaptive soft-argmax (SS) to predict the facial keypoints coordinates from the heatmap.
In order to better evaluate the superiority of SS over $coordinate$ $regression$ and $argmax$,
We conduct different experiments with DRNet.
For $coordinate$ $regression $, we replace the SS by a fully connected layer to directly regress the keypoints coordinates.
We adopt the global average pooling on the scale proposal $\mathcal{S}$ indicated in Sec.~\ref{arch} to convert it to a feature vector with the fixed size.
A fully connected layer with output $2K$ is applied to regress the keypoint coordinates where $K$ means the keypoint number and $L2$ loss is used for optimization.
For $argmax$, each channel in the specific location $\mathcal{S}$ of $M$ corresponding to a specific keypoint and only the coordinates existing keypoints will be set to 1.
The loss function is the same as Eq.~\ref{bceloss}.
In the ablation study, we further conduct FDDB$_{-90^{\circ}}$, FDDB$_{90^{\circ}}$ and PFDDB as the additional benchmarks.
FDDB$_{-90^{\circ}}$ and FDDB$_{90^{\circ}}$ are generated by rotating the FDDB with $-90^{\circ}$ and $90^{\circ}$, respectively.
PFDDB is assigned by all of the images from FDDB which containing profile faces (Roll $>$ $30^{\circ}$ or Yaw $>$ $30^{\circ}$ or Pitch $>$ $30^{\circ}$).

\begin{figure*}[h]
  \centering
  \subfigure[FDDB]{
    \label{fig:subfig:a} 
    \includegraphics[width=1.5in]{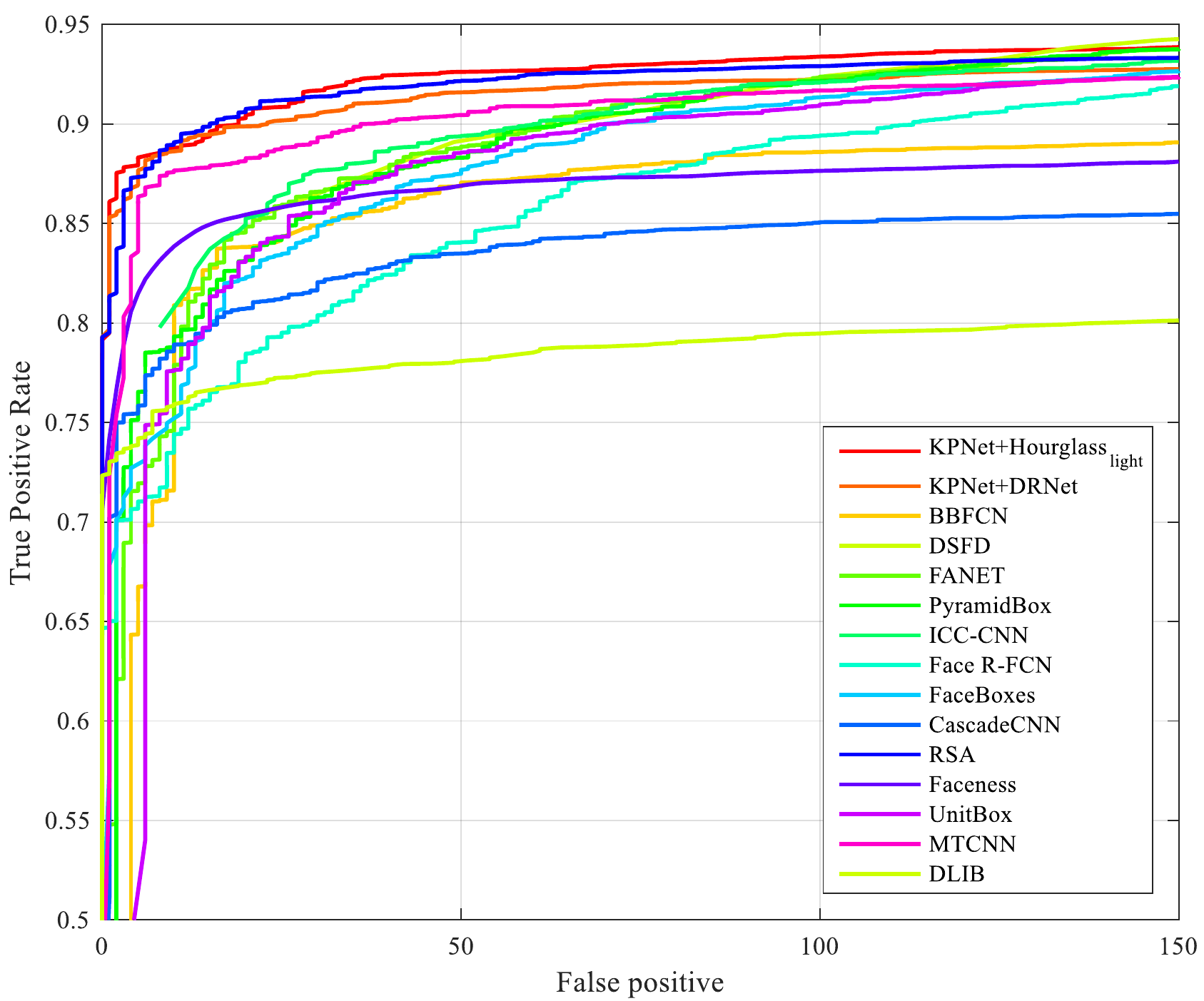}}
  \hspace{0.1in}
       \subfigure[AFW]{
    \label{fig:subfig:c} 
    \includegraphics[width=1.5in]{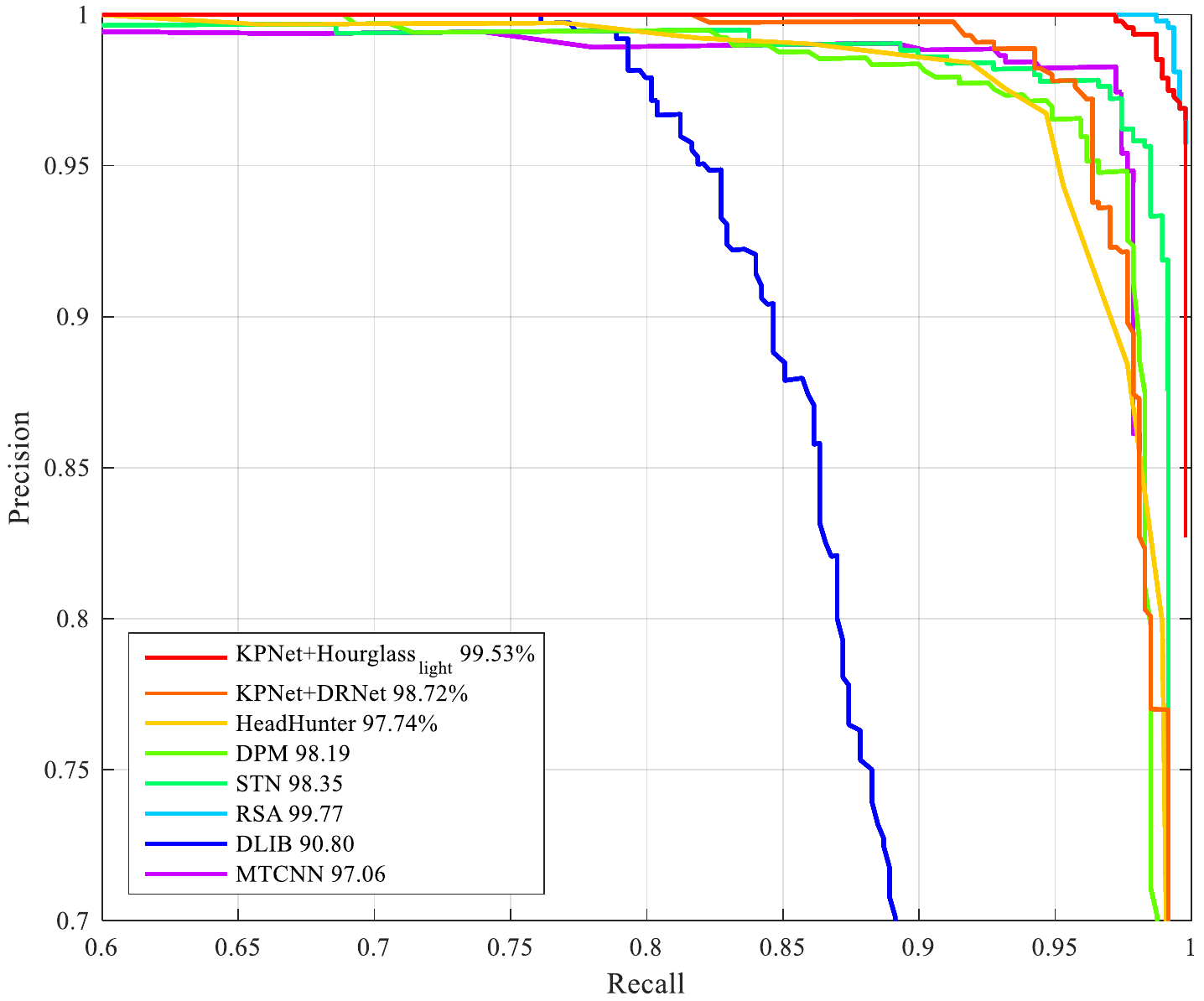}}
    \hspace{0.1in}
  \subfigure[MALF]{
    \label{fig:subfig:d} 
    \includegraphics[width=1.5in]{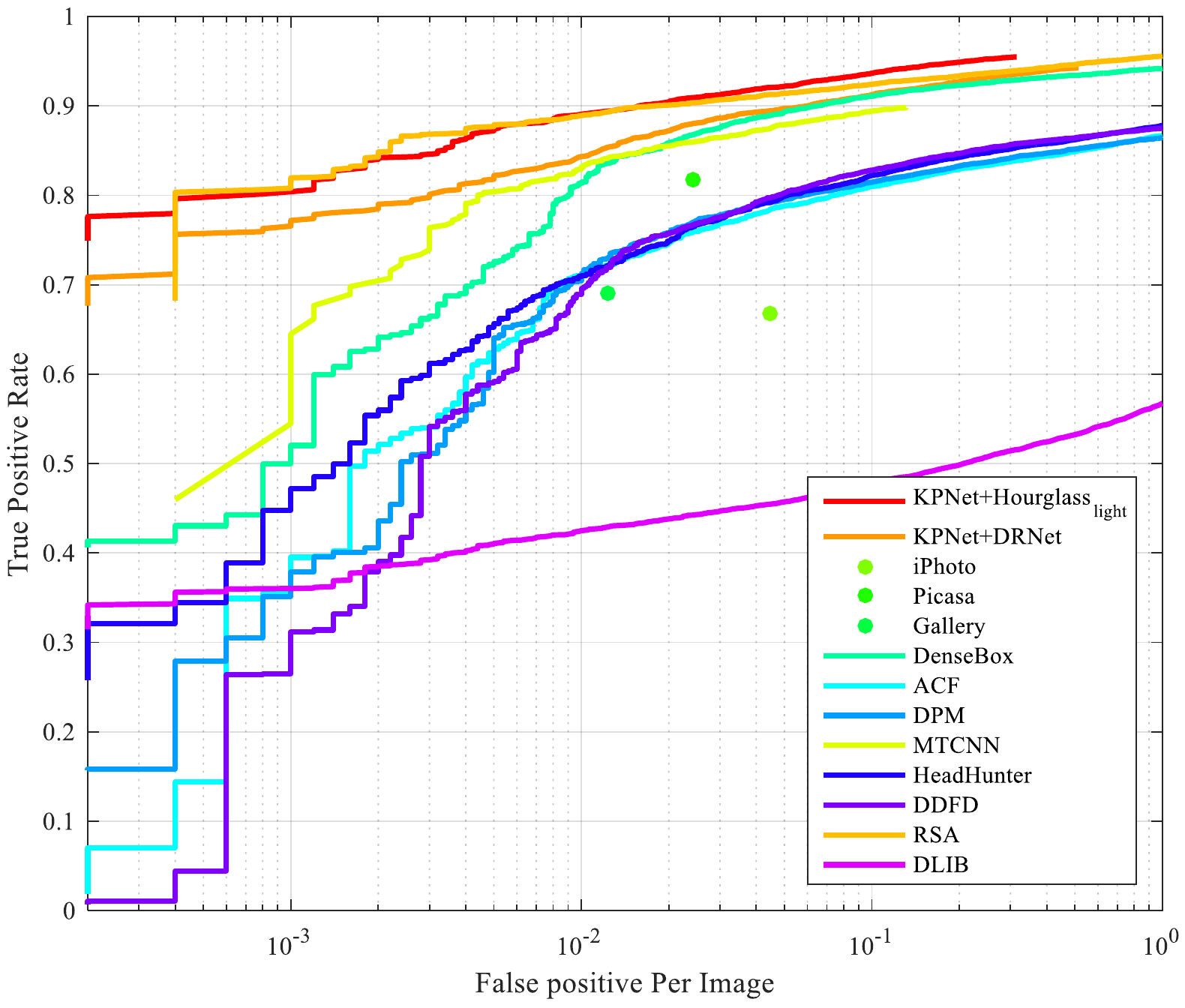}}
  \caption{Comparison to the state-of-the-art on face detection benchmarks.
The proposed KPNet with low-resolution input and lightweight architecture can achieve comparable results with other well-designed anchor-based algorithms.}
  \label{fig:com} 
\end{figure*}

\begin{figure}[t]
\centering
\includegraphics[width=1\linewidth]{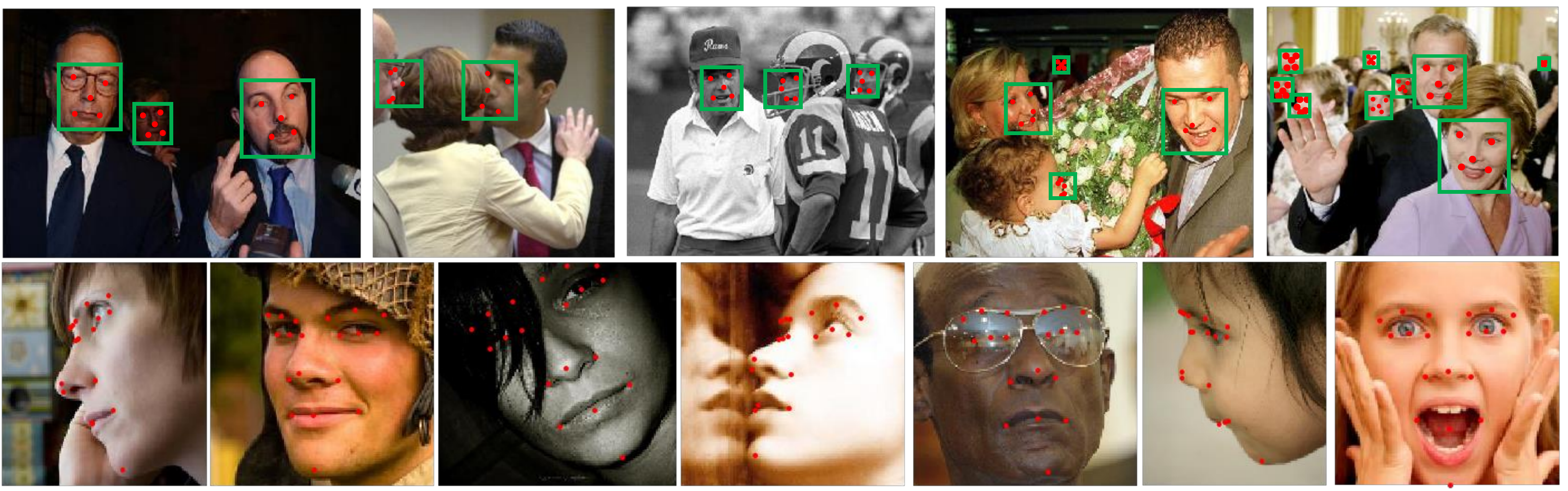}
   \caption{Visualization of joint face detection and alignment. Images are sampled from FDDB and AFLW.}
\label{fig:vis}
\end{figure}
 
 \begin{table}[t]
\centering
\begin{center}
\resizebox{.95\columnwidth}{!}{
\begin{tabular}{c|c|c|c|c}
\hline  
Methods & FDDB& FDDB$_{-90^{\circ}}$ & FDDB$_{90^{\circ}}$ & PFDDB \\
\hline
RSA$_{base}$ & 92.15 & 57.67 &56.78 & 92.8/59.93 \\
$regression$  & 90.86 & 68.96 &67.2 & 90.94/ 47.97\\
$argmax$ & 88.98 & 50.65 &49.9& 83.51/43.9\\
SS &  91.6& 69.32 & 69.97& 91.29/61.44\\
\hline 
\end{tabular}}
\end{center}
\caption{ Performance on different test sets. We report the recall at false positive number 50.
All of the experiments except RSA$_{base}$ are based on DRNet. The two values in PFDDB mean the evaluation on
IOU 0.7 and 0.8, respectively.}
\label{tab:compare_ablation}
\end{table}
  
 \begin{table}[t]
\centering
\begin{center}
\resizebox{.95\columnwidth}{!}{
\begin{tabular}{c c c c |c}
\hline  
Backbone & SP & SS & GT box & FDDB \\
\hline
DRNet & \checkmark & & &  90.7/47.3\\
DRNet & \checkmark & \checkmark & &  91.6/81.1\\
DRNet &  & \checkmark & \checkmark &  96.6/96.6\\
\hline
Hourglass$_{light}$ & \checkmark & & &  91.64/12.52 \\
Hourglass$_{light}$  & \checkmark & \checkmark & &  92.61/80.9\\
Hourglass$_{light}$  &  & \checkmark & \checkmark &  96.72/96.72\\
\hline
\end{tabular}}
\end{center}
\caption{ Ablation study for error analysis.
The recall values are evaluated at false positive number 50 and 1 on FDDB.
SP, SS and GT box mean the scale proposal, scale adaptive soft-argmax and ground-truth face boxes.}
\label{tab:error_ana}
\end{table}
The results are shown in Tab.~\ref{tab:compare_ablation}.
All of the models are trained on normal faces without rotation augmentation.
SS performs better than others, even the anchor-based RSA$_{base}$ with high-resolution input and image pyramid,
KPNet can achieve comparable performance, even more, robust in face-rotation scenarios.
In actually,
detecting face boxes from key points is important
for the lightweight network than directly regressing bounding
boxes. Key points have less uncertain information which
is easier for lightweight models to fit. Besides, the semantic
information of the landmarks is fixed, even if the face
angle/pose changes. This ensures its robustness to face angle/
pose variance.

{\bf{Error analysis.}}
KPNet simultaneously outputs fine-grained scale heatmap and landmarks response map.
To understand how each part contributes to the final error, we perform an error analysis by replacing the predicted scale proposal with the ground-truth boxes.
Furthermore, to evaluate the SS contribution to KPNet, we detect the faces only by scale proposal.

\begin{table}[t]
\centering
\begin{center}
\resizebox{.95\columnwidth}{!}{
\begin{tabular}{c|c} 
\hline  
method & average normalized error \\
\hline
PCD-CNN ~\cite{kumar2018disentangling} & 2.40 \\
TSR~\cite{lv2017deep} & 2.17 \\
LAB~\cite{wu2018look} & 1.25 \\
SAN~\cite{dong2018style} & 1.91\\
PFLD 1X~\cite{guo2019pfld}  & 1.88\\
Wing~\cite{feng2018wing} & 1.65 \\
CPM+SBR~\cite{dong2018supervision} & 2.14\\
\hline
KPNet+DRNet& 1.87 \\
KPNet+Hourglass$_{light}$ & 1.45 \\
\hline 
\end{tabular} 
}
\end{center}
\caption{A comparison of different approaches in terms of the average error normalized ($\times10^{-2}$) on AFLW.}
\label{tab:alignment}
\end{table}

Tab.~\ref{tab:error_ana} shows that the proposed SS can effectively improve the quality of scale proposal and provide the more precise facial location, especially the recall at false positive number 1. Replacing the SP by GT box improves the recall by $4\sim5$\%.
This suggests that there is still ample room for improvement in both SP and SS.

\begin{table*}[t]
\centering
\begin{center}
\resizebox{2.0\columnwidth}{!}{
\begin{tabular}{c c c c c c c c} 
\hline  
Method & Face detection & Face alignment &Online speed &Offline speed& \# Param   & FDDB & AFLW \\
\hline
RSA$_{base}$~\cite{liu2017recurrent} & \checkmark &  & $\sim19.7$ms$_{Ti}$  & W/o &$\sim4$M  & 92.15\% & W/o \\
$S^2AP$~\cite{song2018beyond} & \checkmark &  & $\sim 23.1$ms$_{P100}$ &W/o  &$\sim13.3$M  & 93.5\%& W/o  \\
$S^3FD$~\cite{zhang2017s3fd} & \checkmark &  &$\sim27.8$ms$_{XP}$  & - &$\sim22.46$M  & 92.9\% & W/o\\
MTCNN~\cite{zhang2016joint} & \checkmark &\checkmark  & $\sim31.3$ms$_{Ti}$ &  - & $\sim1.4$M &90.44\%  & 6.9$_*$ \\
DLIB~\cite{king2009dlib} & \checkmark &\checkmark  & $\sim66.7$ms$_{Ti}$ - & - & - & 78.1\% &- \\
PFLD 1X~\cite{guo2019pfld} &  &\checkmark  & $\sim3.5$ms$_{Ti}$ - & - & $\sim3.1$M & W/o & 1.88 \\
LAB~\cite{wu2018look}&  &\checkmark  & $\sim60$ms$_{X}$ - & - & $\sim12.6$M & W/o & 1.25 \\
SAN~\cite{dong2018style}&  &\checkmark  & $\sim343$ms$_{Ti}$ - & - & $\sim199.6$M & W/o & 1.91 \\
Wing~\cite{feng2018wing}&  &\checkmark  & $\sim5.9$ms$_{X}$ - & - & $\sim12.3$M & W/o & 1.65 \\
\hline
KPNet+Hourglass$_{light}$& \checkmark& \checkmark & $\sim10.3$ms$_{Ti}$& $\sim1.6$ms$_{Ti}$ &$\sim1.04$M   & 92.61\% & 1.45 (4.45$_*$) \\
KPNet+DRNet &  \checkmark& \checkmark & $\sim2.6$ms$_{Ti}$ & $\sim1.0$ms$_{Ti}$ &  $\sim1.02$M & 91.6\% & 1.87 (5.77$_*$)\\
\hline
\end{tabular} }
\end{center}
\caption{Comparison with different approaches in terms of the inference speed and performance on FDDB and AFLW.
The online speed means we evaluate it with batch size 1 and the offline speed means we evaluate it with batch size ($>$ 32).
$W/o$ means this application is not supported (e.g. $S^2AP$ can only support the batch size 1 due to its' high-resolution input.).
$Ti$, $P100$, $XP$ and $X$ indicate the GTX 1080Ti, NVIDIA P100, TITAN X Pascal and TITAN X GPU.
The $*$ in MTCNN means this result is normalized by inter-ocular distance and other results on AFLW are normalized by face size.
} 
\label{tab:inference}
\end{table*}
\section{Comparisons with SOTA algorithms}\label{performance}
For face detection, we compare our KPNet with state-of-the-art methods~\cite{liu2019facial,li2018dsfd,zhang2017feature,tang2018pyramidbox,zhang2017detecting,wang2017detecting,zhang2017faceboxes,li2015convolutional,liu2017recurrent,yang2015facial,yu2016unitbox,mathias2014face,chen2016supervised,farfade2015multi,yang2014aggregate,zhang2016joint} and the DLIB c++ library~\cite{king2009dlib}, which supports for joint face detection and alignment.
Fig.~\ref{fig:com} shows the comparison with other approaches on three benchmarks.
On AFW, our algorithm KPNet can achieve 99.53\% AP and 98.72\% AP by Hourglass$_{light}$ and DRNet, respectively.
On FDDB, KPNet+Hourglass$_{light}$ recalls 92.61\% faces with 50 false positives as shown in Fig.~\ref{fig:subfig:a} which outperforms most of the approaches.
On MALF, our methods can also achieve a comparable result with the state-of-the-art.
It should be noticed that the shape and scale definition of the bounding box on each benchmark varies.
KPNet can be easily applied to these benchmarks without complicated design choices.

For face alignment, we compared KPNet with other state-of-the-art methods on AFLW.
AFLW is a challenging dataset that has been widely used for evaluating face alignment algorithms.
As shown in Tab.~\ref{tab:alignment}, our KPNet+Hourglass$_{light}$ outperforms all of the other approaches and KPNet+DRNet can also achieve comparable performance.

\begin{table}[t]
\centering
\begin{center}
\resizebox{.95\columnwidth}{!}{
\begin{tabular}{c|c|c|c|c}
\hline  
Method & FDDB (\%)& AFW (\%) & MALF (\%)& AFLW\\
\hline
RPN+DRNet & 75.82 & 88.82 & 58.68 & 2.12\\
AE+DRNet & 46.96 & 76.8 & 32.57 &2.12\\
\hline 
KPNet+DRNet& 91.6 & 98.72 & 88.92 & 1.87 \\
\hline
\end{tabular} }
\end{center}
\caption{ Comparison with top-down method and bottom-up methods AE proposed in pose estimation.}
\label{tab:anchor}
\end{table}

Furthermore, we compare KPNet+DRNet with the bottom-up associative embedding (AE)~\cite{newell2017associative,law2018cornernet} and top-down RPN+DRNet.
As shown in Tab~\ref{tab:anchor}, 
whether adopting anchor-based RPN or AE, the performance is strictly limited by the capacity of lightweight backbone.
It's in stark contrast that KPNet with fine-grained scale approximation and scale adaptive soft-argmax achieve excellent performance.
To better understand the performance of KPNet, we visualize some images sampled from FDDB and AFLW in Fig.~\ref{fig:vis}.

\section{Analysis of the inference speed}

In this section, we explore the performance and speed in detail compared with other approaches as shown in Tab.~\ref{tab:inference}.
We report the recall at false positive number 50 on FDDB and the NME ($\times10^{-2}$) on AFLW.

In the offline applications, KPNet with DRNet can achieve $\sim$1000 fps at GTX 1080Ti, faster than other face detectors with a large margin.
Even compared with the state-of-the-art algorithms on face alignment, KPNet still has a faster model inference speed.

Both the MTCNN and DLIB are the popular frameworks for joint face detection and alignment.
KPNet outperforms them with a large margin in terms of inference speed and performance.
No complex hyperparameter designing is required so that it can be easily applied to different scenarios.

\section{Conclusion}
This paper proposes a simple, lightweight but accurate framework KPNet which does away with anchor boxes.
It focuses on joint generic face ($>$ 20px) detection and alignment.
Unlike most face detection methods and top-down joint face detection and alignment methods, KPNet adopts the bottom-up mechanism.
It first predicts the facial landmarks from a low-resolution image via the well-designed fine-grained scale approximation and scale adaptive soft-argmax operator. Finally, the precise face bounding boxes, no matter how we define it, can be inferred from the landmarks.
KPNet can effectively alleviate the vague definition of the face bounding box. 
Without bells and whistles, KPNet achieves state-of-the-art accuracy on generic face detection and alignment benchmarks with only $\sim1$M parameters.
The model inference speed can achieve $\sim1000$fps on GPU and it's easily deployed to most modern front-end chips.

{\small
\bibliographystyle{aaai}
\bibliography{egbib}
}
\end{document}